\documentclass[review]{elsarticle}

\usepackage{lineno,hyperref}
\modulolinenumbers[5]

\journal{Applied Ocean Research}

\usepackage{float}

\begin{document}

\begin{frontmatter}

\title{Response Component Analysis for Sea State Estimation Using Artificial Neural Networks and Vessel Response Spectral Data}

%% or include affiliations in footnotes:
\author[unsw_address,dst_address_vic]{Nathan K. Long} %\corref{mycorrespondingauthor}}
% \cortext[mycorrespondingauthor]{Corresponding author}
% \ead{nathan.k.long.91@gmail.com}

\author[dst_address_vic]{Daniel Sgarioto}
\author[unsw_address]{Matthew Garratt}
\author[dst_address_sa,flinders_address]{Karl Sammut}

\address[unsw_address]{University of New South Wales, Campbell ACT, Australia}
\address[dst_address_vic]{Defence Science and Technology Group, Fishermans Bend VIC, Australia}
\address[dst_address_sa]{Defence Science and Technology Group, Edinburgh SA, Australia}
\address[flinders_address]{Flinders University, Tonsley SA, Australia}

\begin{abstract}

The use of the `ship as a wave buoy analogy' (SAWB) provides a novel means to estimate sea states, where relationships are established between causal wave properties and vessel motion response information. This study focuses on a model-free machine learning approach to SAWB-based sea state estimation (SSE), using neural networks (NNs) to map vessel response spectral data to statistical wave properties for a small uninhabited surface vessel.

Results showed a strong correlation between heave responses and significant wave height estimates, whilst the accuracy of mean wave period and wave heading predictions were observed to improve considerably when data from multiple vessel degrees of freedom (DOFs) was utilized. Overall, 3-DOF (heave, pitch and roll) NNs for SSE were shown to perform well when compared to existing SSE approaches that use similar simulation setups. One advantage of using small vessels for SAWB was shown as SSE accuracy was reasonable even when motion responses were low (in high-frequency, low wave height sea states). Given the information-dense statistical representation of vessel motion responses in spectral form, as well as the ability of NNs to effectively model complex relationships between variables, the designed SSE method shows promise for future adaptation to mobile SSE systems using the SAWB approach.

\end{abstract}


\end{frontmatter}

\section{Introduction}

The safety and efficiency of seagoing vessels are greatly affected by operational conditions~\cite{Fossen2011_ch8}. Due to the relationship between vessel motion and the wave properties that cause them, the ability to estimate wave conditions, such as wave heights, frequencies, and directions, offers a number of major advantages to operators of maritime platforms.

A vessel can modify its route, based on sea state information, to reduce the likelihood of extreme motions and impact loads on the ship. This can help to increase the comfort and safety of crew and passengers onboard, as well as protect cargo and vessel systems from unwanted damage. Further, vessel route optimisation can reduce fuel consumption by traversing calmer waters. For example, by using a dynamic sea state estimation (SSE) system, regions of rough seas could be avoided entirely by investigating the state of the sea ahead of a maritime platform.

One emerging solution to the SSE problem is the use of the `ship as a wave buoy' (SAWB) analogy. The SAWB technique involves establishing the relationship between a vessel\textquoteright s motion and the causal wave properties~\cite{Nielsen2017}. As such, the vessel can be deployed in the sea and its motion responses recorded, then interpreted to estimate the encountered wave conditions. 

Previous studies have considered large ships as their models for SAWB (eg.~\cite{Mak2019,Mak2019_2,Kawai2021,Mittendorf2022}), where the vessels would be unlikely to digress from global shipping routes. While knowledge of sea states along shipping lanes is important, there are a wide variety of environments where sea state information could prove vital to operational success. The use of uninhabited surface vessels (USVs) for SSE creates a platform which can be deployed in-situ and in real-time, as well as being dynamically operable. A major advantage of using small USVs is that they can be employed in large numbers across large areas, including shallow water, and are not confined to predefined shipping routes. Further, USV platforms are relatively cheap and disposable, making them ideal for rapid deployment in uncertain or volatile environments. As such, investigation of the performance of small USVs using SAWB for SSE is critical for obtaining a more comprehensive understanding of the practicality of the SSE technique.

Researchers often look to nature when solving complex problems or looking for ways to augment existing processes, such as by using biologically inspired machine learning (ML) algorithms. Artificial neural networks (NNs) are simplified representations of the neural connections in the brain, which can be used to map complex relationships between input data and output variables~\cite{Jain1996}.

This paper presents a model-free approach, based on NNs, to define the relationship between sea states and vessel motions. The NNs were developed to take vessel spectral response data as input, then output wave properties. Specifically, heave, pitch, and roll data was used as input, while significant wave height, mean wave period, and wave heading represent the outputs.

While most other ML-based SSE techniques using SAWB have used raw time-series response data as input~\cite{Mak2019,Mak2019_2,Cheng2020}, the use of spectral response information provides an information dense alternative~\cite{Kawai2021,Han2022,Mittendorf2022}, with potential for future use in online SSE. The simplicity of the proposed SSE methodology was designed such that it could be quickly and easily applied to any vessel of opportunity.

As such, 3-DOF vessel response time-series data was simulated for a generic small USV for a range of parametric Modified Pierson-Moskowitz (MPM) sea states, then transformed into their power spectral density (PSD) representations to use as input to the designed NNs. Additionally, the total power and forward speed were recorded to accompany the PSD response data to train the SSE models.

Furthermore, a study was conducted into the estimation performance when each DOF response component was used separately, as well as combined 2-DOF models and 3-DOF models, to investigate the relative contributions information from each DOF and various combination makes to the NN-based SSE method. As such, 21 separate NNs for SSE were developed, one for each estimated wave property for the single DOF, combined 2-DOF, and 3-DOF models. 

Given that ships act like low-pass filters~\cite{Nielsen2007}, larger vessels are less responsive to high-frequency wave components. Many other studies on using ML for SAWB SSE have used large vessels in their models~\cite{Mak2019,Mak2019_2,Kawai2021,Mittendorf2022}. This study aims to investigate the SSE performance of small craft when traversing high-frequency and low wave height sea states, given their relatively higher motions in response to high frequency wave components.

This paper provides a review of background literature on SSE in Section~\ref{sec:back}, the simulation models, response data transformation, and NN construction methodologies are given in Section~\ref{sec:meth}, results and discussion are presented in Section~\ref{sec:res}, before concluding remarks are made in Section~\ref{sec:concl}.

\section{Background} \label{sec:back}

Several different techniques have been developed to estimate sea states, with some of the most common methods including the use of wave buoys~\cite{NRC1998}, ship-based radar~\cite{Huang2017}, satellite-based remote sensing~\cite{Yu2014}, and an emerging method of using the `ship as a wave buoy analogy'~\cite{Nielsen2017}.

The most common method of taking sea state measurements is via wave buoys, floating sensor-mounted platforms~\cite{Cardone2014}. However, they are few and far apart~\cite{JCOMMOPS2019}, and contain no means of locomotion, thus making them impractical for dynamic operability as part of in-situ applications. Ship-based radar systems solve this problem by allowing measurements to be taken during ocean transit; however, the radar requires frequent calibration, is expensive to install, and interpretation of the radar data is computationally demanding~\cite{Brodtkorb2015}. 

Satellite-based altimetry encompasses the procedures which involve measuring the time taken for a radar pulse, transmitted from satellite, to reflect from the sea surface and back to the satellite or a separate receiver. The spatial resolution of satellite altimetry, however, is generally quite crude (on the order of $10-100km$), as well as having coarse temporal resolutions~\cite{Maqueda2016}. Global navigation satellite system (GNSS) geodesy is the triangulation between a GNSS receiver at the ocean surface and global navigation satellites. GNSS geodesy is capable of measuring displacements on the scale of decametres to centimetres~\cite{Maqueda2016,Maqueda2017,Penna2018}. However, its overall complexity and additional noise source from a ship\textquoteright s responses to wave excitation make it less appealing for simple in-situ SSE while at sea, when compared to the SAWB method. 

The SAWB analogy applies the principles of wave buoys to ocean-going vessels, where the relationship between a vessel\textquoteright s motion and causal wave properties are modeled. The vessel can then be placed in an unknown sea state and use its motion response information to estimate the causal wave conditions~\cite{Nielsen2017}. SAWB shows promise due to its potential for relatively simple, in-situ, near real-time SSEs. While all of the SAWB methods interpret the vessel responses to wave excitation, there are a few key defining features that delineate the methodologies. One major distinction in SAWB is between vessel model-based techniques and model-free approaches.

In model-based SAWB approaches, vessel-dependent transfer functions describe how the wave motion is transferred to vessel responses. Transfer functions can be calculated using three-dimensional (3D) panel codes based on potential wave theory, strip theory, and computational fluid dynamics programs~\cite{Nielsen2017}. One of the main assumptions of transfer functions is that moderate wave heights form a linear relationship with the motion responses. However, this assumption can provide unreliable results in severe wave conditions, where the linear relationship no longer holds, as well as for non-conventional hullforms~\cite{Nielsen2019}.

Model-free approaches are designed independently to the vessel parameters, instead defining a relationship between input and output data alone. In other words, a connection between known vessel responses to known wave input is created without passing through a physics-based model. Therefore, model-free approaches rely only on data interpretation, usually via ML or a type of filter, circumventing the development of complex transfer functions.

    \subsection{ML-augmented SSE}

    The augmentation of SSE using SAWB via the application of ML algorithms has recently gained attention. ML techniques used include quadratic discriminant analysis and partial least square regression~\cite{Arneson2019}, as well as deep learning~\cite{Mak2019,Mak2019_2,Cheng2020,Cheng2020_2,Kawai2021,Han2022,Mittendorf2022}. 

    Arneson, Brodtkorb, and S{\o}rensen~\cite{Arneson2019} separated the SSE problem into wave heading categorization, and wave height and frequency estimation. Wave headings were discretized, representing a classification problem, where quadratic discriminant analysis (QDA) was used to estimate the wave heading classification. Partial least squares regression (PLSR) was then performed for each heading classification to estimate wave heights and periods. The authors transformed vessel response time-series data into frequency-domain response spectra, then took the total power of the spectra to use as input to the QDA and PLSR models. While the authors described their approach as underperforming when compared with model-based approaches, its ability to estimate both low and high sea states highlights one of the major benefits of model-free SAWB. However, fluctuations in the results indicate that further modifications are necessary to improve the robustness of the techniques. 

    Wave direction is cited as being the most difficult parameter to accurately estimate when using traditional wave estimation methods~\cite{Mak2019}. Therefore, Mak and D\"{u}z~\cite{Mak2019} focused on improving directional wave estimates by training three different types of convolutional neural networks (CNNs) to estimate wave headings using ship response data: a CNN for multivariate regression (CNN-REG), a long short-term memory recurrent neural network (MLSTM-RNN) CNN, and a Sliding Puzzle CNN. The authors used 6-DOF time-series response data as input to the CNNs, where data that was input in chronological order was compared to shuffled data, following 5-fold cross-validation. Results showed lower error for the shuffled data, while the MLSTM-RNN CNN and Sliding Puzzle CNN generally outperformed the CNN-REG.
        
    Mak and D\"{u}z~\cite{Mak2019_2} then tested the performance of their NNs on simulated data. The simulations were conducted using three different ship types, where the same wave input was simulated for each ship in a sea state represented using JONSWAP spectra (a function of significant wave height and mean wave period). The Sliding Puzzle and MLSTM-RNN CNNs were able to estimate the headings with an approximate $1^\circ$ error for 95$\%$ of results, while the CNN-REG was able to estimate the heading with an approximate $4^\circ$ error, for all ship types, indicating high performance for NN-based SAWB approaches.
        
    A study by Cheng et al.~\cite{Cheng2020} developed a densely connected CNN (dubbed SSENET) to estimate wave heights and directions using ship response data for a vessel with forward speed. Cheng et al.\textquoteright s method is unique as they introduced a zigzag shaped path for the vessel to follow when collecting response data. The authors grouped the sea states into five wave height categories and separated the wave direction into eight categories, resulting in 40 different sea state combinations. 9-DOF time-series vessel response measurements were used to train the SSENET, taken from 12 real data sets, as well as two simulated data sets. SSENET was able to achieve high accuracy on most of the real data sets, reaching a maximum of 99.61$\%$ accuracy when classifying one of them, and an average accuracy of 82.99$\%$ for all 12 real data sets. SSENET, further, achieved an 89.81$\%$ and 93.46$\%$ accuracy for the two simulated data sets. When compared to other NN types, the SSENET results showed the highest accuracy.
        
    Another study by Cheng et al.~\cite{Cheng2020_2} used vessel response spectrograms as input to the CNNs instead of raw time-series data (SpectralSeaNet). Spectrograms describe a combination of both time and frequency information as two dimensional (2D) images. The SpectralSeaNet results were found to outperform a CNN and LSTM trained on raw time-series data, where SpectralSeaNet had an average accuracy of 94$\%$, while the raw time-series CNN had 91$\%$ and the LSTM had $79\%$ average accuracy respectively. This study indicates that use of spectral data for NNs designed for SSE shows promise.
    
    Kawai et al.~\cite{Kawai2021} designed a CNN to estimate parametric directional wave spectra. Simulated heave, pitch, and roll response spectra, as well as vertical bending stress spectra and corresponding cross-spectra, of a 14,000TEU container ship were used as input. The CNN then estimated four parameters to describe the Ochi–Hubble directional sea states. The estimation accuracy for significant wave height, modal wave frequency, and kurtosis were high, with error standard deviations of $0.106m$, $0.015rad/s$, and 0.382, respectively. Wave direction estimation had a much greater spread of errors, with a standard deviation of $20.93^\circ$. Estimation accuracy was found to reduce when total power in the vessel responses were low, which could be caused by the reduced motion of large ships to high-frequency wave components.
    
    An alternative approach to directional wave spectrum estimation was developed by Han et al.~\cite{Han2022}, where they used a generative adversarial neural network (GANN) to estimate a 2D wave spectrum. The GANN was comprised of two CNNs, one was used to estimate the spectrum, then the second was used to discriminate whether the estimated spectrum was realistic or not. Nine vessel motion cross-spectra were simulated using a R/V Gunnerus vessel model for sway, heave, and pitch responses to double-peak Pierson-Moskowitz sea states to train the GANN. While a 2D wave spectrum was estimated by the GANN, the significant wave height, mean wave period, mean wave direction, and directional spreading were calculated to compare the GANN to Bayesian and non-adversarial NN approaches. Results found that the GANN estimations were more accurate than either the Bayesian or non-adversarial NN estimations on average, except the Bayesian method had higher accuracy mean wave direction estimations. While comprehensive, the authors did not investigate the effects of vessel forward speed on estimation accuracy in their paper.
    
    Finally, a paper by Mittendorf et al.~\cite{Mittendorf2022} conducted a comparative study between time-based and frequency-based ML approaches to SSE using recorded ship motion data for a 2800 TEU Panamax container vessel. Four different NN architectures were investigated to estimate significant wave height, peak wave period, and mean wave encounter direction, where the Inception NN output the highest accuracy estimations. A comparison between a multi-output regressor (MOR) NN model approach and a multi-task learning (MTL) approach was also undertaken, where the MTL models were trained for parallel transfer learning, such that the estimated wave parameters each had a separate output layer from the network rather than all being output by the same layer. The MTL approach was found to generally have higher estimation performance than the MOR approach, particularly in the time domain. Overall, the results revealed that the frequency-domain approach to SAWB SSE estimated wave properties with higher accuracy, as well as being more computational efficient than the time-based approach.
    
    While literature on the use of ML for SSE using the SAWB technique is growing, further investigation of different methodologies should be investigated for different scenarios and operations to validate and expand its practical application. This study focuses on the use of small USVs for SAWB, with a simple methodology which could be applied to experimental data with a simple, robust methodology that is applicable to a wide range of vessels of opportunity.
    
    Further, it is hypothesized that small vessels will estimate sea states with higher accuracy when subjected to lower power motion responses, when compared to the larger vessels previously considered~\cite{Kawai2021}.

\section{Methodology}\label{sec:meth}

    \subsection{Wave Modeling}
    
    The stochastic nature of the sea surface can be approximated as a composition of multiple sinusoidal waves with varying parameters. Equation~\ref{eq:wave_sum_z_w} defines wave elevation ($\zeta$) as a function of time ($t$) and uni-directional wave direction ($x_w$), where $\omega$ represents the wave frequency, $A_w$ is the wave amplitude, $k_w$ is the wavenumber, and $\epsilon_w$ is the random phase shift of each of the $N_w$ wave components. The direction of wave propagation is considered relative to north-east-down coordinates. 
    \begin{equation} \label{eq:wave_sum_z_w}
        \zeta (x_w,t) = \sum_{n=1}^{N_w} A_{w_n} \ \sin(k_{w_n} x_w - \omega_{n} t + \epsilon_{w_n})
    \end{equation}
    
    Sea states are often characterized by their statistical properties, where three of the most commonly used properties are significant wave height ($H_s$), mean wave period ($T_1$), and mean wave heading~\cite{Wang2016}. Significant wave height represents the average height of the tallest one-third of wave heights, while the mean wave period is the average period, and mean wave heading is the average heading, of all waves measured.
    
    A number of idealised wave spectra have been developed for different wave conditions in order to model the sea surface. The Modified Pierson-Moskowitz (MPM) spectrum was developed to model a fully developed sea~\cite{Fossen2011_ch8}, parameterized using the significant wave height and zero-crossing wave period ($T_z$), given in Equations~\ref{eq:mpm_spec} to~\ref{eq:mpm_spec_bpm}. The zero-crossing wave period can be calculated using the mean period as $T_z = 0.9212 T_1$~\cite{Lantos2010}.
    \begin{equation} \label{eq:mpm_spec}
       S_{\zeta_{PM}}(\omega) = \frac{A_{PM}}{\omega^5} \ \exp\left(\frac{-B_{PM}}{\omega^4}\right)
    \end{equation}
    \begin{equation} \label{eq:mpm_spec_apm}
       A_{PM} = \frac{H_s^2}{4 \pi} \left( \frac{2 \pi}{T_z} \right)^4
    \end{equation}
    \begin{equation} \label{eq:mpm_spec_bpm}
       B_{PM} = \frac{1}{\pi} \left( \frac{2 \pi}{T_z} \right)^4
    \end{equation}

    The wave elevation time-series can then be calculated from a wave spectrum by randomizing the phase components and substituting the wave amplitudes in Equation~\ref{eq:wave_sum_z_w} with those calculated in Equation~\ref{eq:spec_ord_amp}, where $S_\zeta$ is the spectral ordinate and $d \omega$ represents the discretized frequency components.
    \begin{equation} \label{eq:spec_ord_amp}
        A_{w_n} = \sqrt{2 S_{\zeta_n}(\omega_{n}) \ d\omega}
    \end{equation}
    
    \subsection{Ship Model}
    
    It was decided to simulate the motion response of a generic small USV to varying sea states generated using MPM spectra in order to estimate wave properties. The vessel properties used in the simulated model are given in Table~\ref{tab:USV}.
    
    \begin{table}[H]
		\begin{center}
			\caption{Small USV model parameters.}
			\label{tab:USV}
			\begin{tabular}{l | c}
				Parameter	& Value \\ \hline \hline
				% Mass ($m_\pi$) & 55$kg$ \\
				Length ($L_\pi$) & 2.05$m$  \\
				Breadth ($B_\pi$) & 0.61$m$ \\
				Draught ($T_\pi$) & 0.16$m$ \\
				Water plane coefficient ($C_{wp_\pi}$) & 0.877 \\
				Block coefficient ($C_{b_\pi}$) & 0.731 \\
				Displacement ($\Delta_\pi$) & 150$kg$ \\
		        Transverse metacentric height ($GM_{T_\pi}$) & 0.264$m$
			\end{tabular}
		\end{center}
	\end{table}

    \subsection{Vessel Transfer Functions}
    
    A vessel has six DOFs related to the motion of its entire body, three translational and three rotational. While 6-DOF motion has been used before to estimate sea states~\cite{Arneson2019,Mak2019,Mak2019_2}, the use of 3-DOFs have been the most common: heave ($z_\pi$), pitch ($\theta_\pi$), and roll ($\phi_\pi$)~\cite{Nielsen2006,Hofler2017,Hinostroza2016,Nielsen2019}. Therefore, frequency response functions (FRFs), which are frequency-wise transfers functions that translate wave motion to vessel motion, have been used to calculate heave, pitch and roll motion responses. However, horizontal displacements have been assessed in order to relate the vessel spatially to the wave system coordinates by considering the vessel\textquoteright s forward velocity. 
    
    While a vessel\textquoteright s motion for zero forward speed can be calculated directly using the FRFs, when it has a velocity, the motion excitation frequency becomes a function of the wave encounter frequency ($\omega_e$). The wave encounter frequency is a function of vessel speed ($U_\pi$) and relative wave heading ($\mu_h$), as calculated in Equation~\ref{eq:enc_freq}.
    \begin{equation} \label{eq:enc_freq}
        \omega_e = \omega - k_w U_\pi cos(\mu_h) \equiv \alpha_\Phi \omega
    \end{equation}
    
    The relative wave heading describes the angle between the body-fixed axis of a vessel and the wave-relative direction of wave propagation.
    
    In order to simulate the heave, pitch, and roll motion of the USV to wave excitation, it was decided to use the semi-analytical approach developed by Jensen, Mansour, and Olsen~\cite{Jensen2004}. The series of closed-form expressions developed by~\cite{Jensen2004} only require the following vessel-specific information: vessel length, breadth, draught, displacement, block coefficient, water plane area, and transverse metacentric height (given in Table~\ref{tab:USV}). The vessel is considered as rectangular prism(s) with homogeneous properties. Previous SSE studies using the SAWB approach that have simulated ship responses using the equations derived in~\cite{Jensen2004} include~\cite{Nielsen2016,Brodtkorb2018,Nielsen2018,Nielsen2008}. 
    
    Equations~\ref{eq:heave_eom} and~\ref{eq:pitch_eom} are the equations of motion (EOMs) for pitch and heave, which define the basis for their FRFs. The forces from wave excitation are represented as $F_0$ for heave and $G_0$ for pitch, while the dimensionless sectional hydrodynamic damping coefficient is shown as $Q_\pi$.
    \begin{equation} \label{eq:heave_eom}
	    \left(\frac{2 k_w T_\pi}{\omega^2}\right) \ddot{z}_\pi + \left(\frac{Q_{\pi}^2}{k_w B_\pi \alpha_\Phi^3 \omega}\right) \dot{z}_\pi + z_\pi = A_w F_{0_z} \ cos(\omega_e t)
    \end{equation}
    \begin{equation} \label{eq:pitch_eom}
	    \left(\frac{2 k_w T_\pi}{\omega^2}\right) \ddot{\theta}_\pi + \left(\frac{Q_{\pi}^2}{k_w B_\pi \alpha_\Phi^3 \omega}\right) \dot{\theta}_\pi + \theta_\pi = A_w G_{0_\theta} \ sin(\omega_e t)
    \end{equation}
    
    In order to calculate the EOM for roll, the hydrodynamic damping coefficient of the vessel ($B_{\pi_{44}}$) is found by assuming that the vessel is comprised of two prismatic sections of equal draught, but with different breadths and corresponding cross-sectional areas. As such, the vessel model is more complex for roll than heave and pitch, which is required to capture the desired roll characteristics. 
    
    Equation~\ref{eq:roll_eom} gives the EOM for the roll response, where $T_{N_\phi}$ represents the natural period for roll (from~\cite{Kruger2008}), $C_{\pi_{44}}$ is the restoring moment coefficient, and $M_\phi$ is the roll excitation moment.
    \begin{equation} \label{eq:roll_eom}
	    \left(\frac{T_{N_\phi}}{2\pi}\right)^2 C_{\pi_{44}} \ddot{\phi} + B_{\pi_{44}} \dot{\phi} + C_{\pi_{44}} \phi = M_\phi
    \end{equation}
    
    Jensen, Mansour, and Olsen\textquoteright s FRFs provide a simplified representation of the complex interactions between monohull vessels and encountered waves. However, as the vessel response DOFs are assumed to be decoupled, information about DOF cross-spectra information cannot be extracted. Therefore, due to vessel lateral-plane symmetry, differentiation between port and starboard wave headings cannot be made. This relatively simple modelling approach was used primarily for convenience to demonstrate the proposed NN-based SWAB SSE technique. In reality, the methodology would be applied in-situ at sea on a vessel of opportunity to recover sea state characteristics from measured motion responses. This is the subject of on-going investigations and will be reported separately in a future publication.
    
    The full derivations of Equations~\ref{eq:heave_eom} to~\ref{eq:roll_eom} can be found in~\cite{Jensen2004}.

    \subsection{Response Spectra}
    
    While other NN-based SAWB SSE techniques used raw time-series data as input, it was hypothesised that response features could be manually extraction by transforming the response time-series data into their PSD representations~\cite{Deng2012}. PSDs describe how power in the time-series data is distributed across response frequency components, and can be estimated via discrete Fourier transformation (DFT)~\cite{Solomon1991}.
    
    However, as discussed by Nielsen~\cite{Nielsen2006}, a type of smoothing function is desirable when transforming vessel response time-series data into the frequency domain due to the volatility of the derived spectral magnitudes resulting from a DFT. As such, it was decided to use Welch\textquoteright s method to estimate the PSD of the vessel response time-series. An in-depth description of Welch\textquoteright s method is given in~\cite{Solomon1991}.
    
    Following experimentation, the \textit{Hann} window was chosen for this analysis, where 15 windows were used for heave and roll transformations, and 13 were used for the pitch transformation. Likewise, the same 50$\%$ overlap between windows was selected for heave and roll, while 80$\%$ was used for pitch. 
        
    The spectral ordinates with the greatest values and their corresponding frequency components were then extracted from the PSDs to be used as input for the SSE NN algorithms. Furthermore, in a similar manner to the study by Arneson, Brodtkorb, and S{\o}rensen~\cite{Arneson2019}, the PSD was integrated to obtain the total power of the response spectra to be used as input to the NNs.
    
\begin{figure}[H]
        \centering
        \includegraphics[scale=0.8]{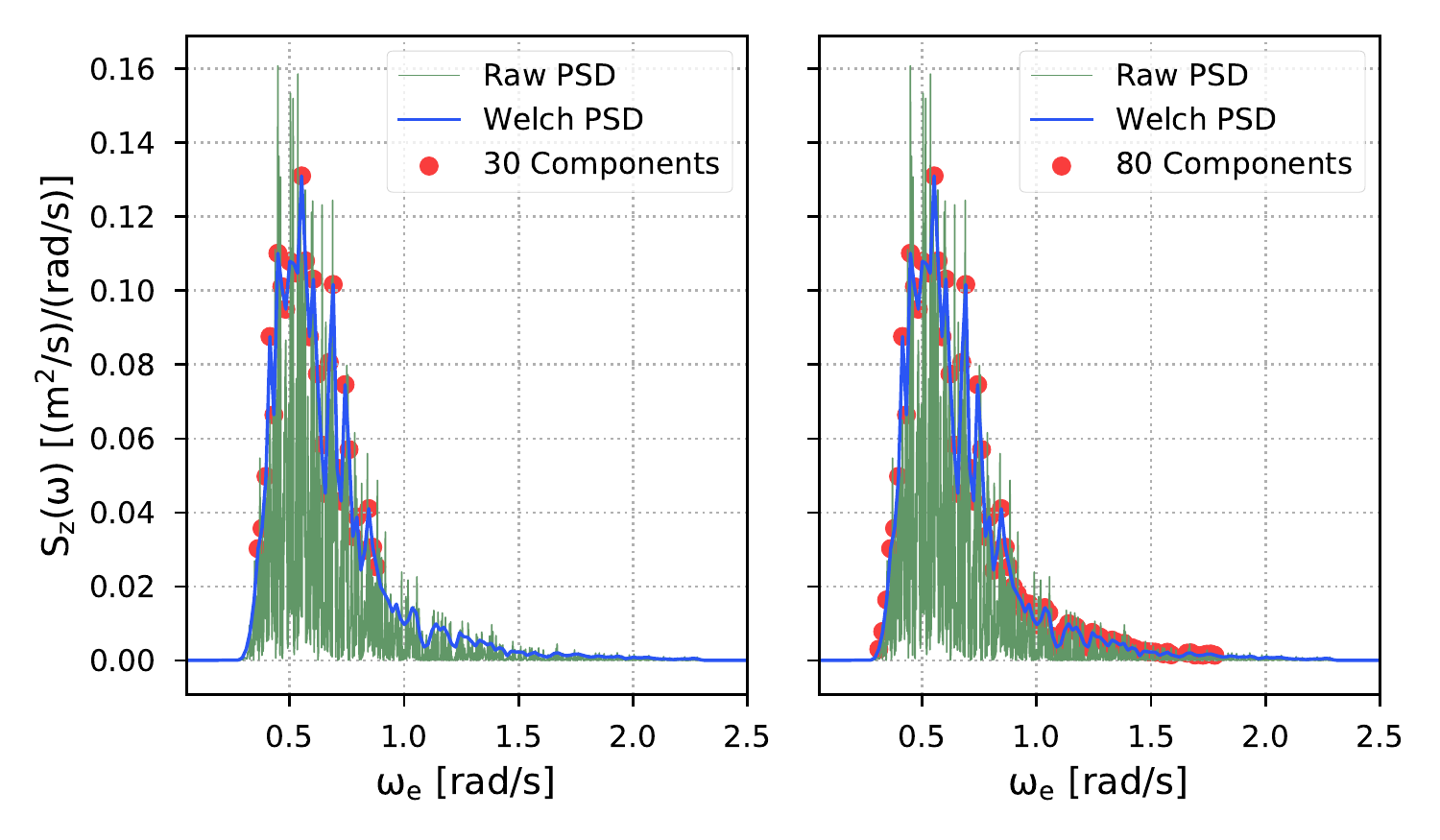}
        \caption{Heave response PSD ($U_\pi = 2m/s$, $\mu_h = 130^\circ$, $H_s = 1.2m$, $T_1 = 8.5$).}
        \label{fig:z_welch}
    \end{figure}

    Figure~\ref{fig:z_welch} gives an example of a PSD for a heave response time-series for 40 minutes. The raw PSD is shown as the thinner green line, the Welch approximation is given as the thicker blue line, and the highest energy components are highlighted as red markers. The left plots give an example of the 30 PSD components using the Welch method with the highest energy, while the right plots show the 80 components with the highest energy. 
    
    The raw PSD is highly volatile, making feature extraction more difficult, while the smoothed PSD derived by Welch\textquoteright s method provides a good overall approximation of the power distribution. It is hypothesised that the components with the greatest energy of the PSD constitute the most important response information, providing feature-extraction prior to SSE NN training. This reduces the overall complexity required for the NNs, as well as reducing the computational time to train and use the networks, when compared with including all spectral ordinates from the PSDs.

    \subsection{Sea State Estimation}
    
    In order to estimate sea states, vessel motion responses were analysed for different sea and vessel states so that they could be used as input for artificial NNs, which were designed to output the significant wave height, mean wave period, and relative wave heading. 
        
    In order to build the NN for SSE, a significant amount of training data was required. Therefore, it was decided to run 48000 simulations of 40 minutes (2400 seconds), broken into 8192 time steps. 
        
    For each simulation, an MPM sea spectrum was generated and transformed into a wave system in time and space, where the spectra were comprised of 500 wave frequencies (as in~\cite{Brodtkorb2018}) ranging from $0.05rad/s$ to $2.0rad/s$, then combined with the FRFs from~\cite{Jensen2004} to produce the vessel response time-series.
    
    Table~\ref{tab:sim_vars} defines the range of each variable tested, which were sampled randomly from uniform distributions.
    
    \begin{table}[H]
	    \begin{center}
			\caption{Simulation variable ranges}
			\label{tab:sim_vars}
			\begin{tabular}{l | c}
				Variable	& Range \\ \hline \hline
				Significant wave height & $0.5m$ to $2.5m$ \\
				Mean wave period & $4s$ to $13s$ \\
				Relative wave heading & $0^\circ$ to $180^\circ$ \\
				Vessel speed & $0m/s$ to $5m/s$ \\
			\end{tabular}
		\end{center}
	\end{table}
    
    It was decided to add noise to the time-series responses to better approximate response measurements using platform onboard sensors. The standard deviations of the noise present in the inertial measurement unit of a waverider buoy~\cite{Gryazin2016} were given as $0.01m$ for heave, 0.028$^\circ$ for pitch, and 0.011$^\circ$ for roll. These values were used to generate random noise for each step of the response time-series (assuming a Gaussian distribution for the noise).

	    \subsubsection{Neural Network Architecture}
	    
	    From experimentation, it was determined that 30 PSD components provided sufficient information to estimate significant wave height and mean wave period well, while 80 PSD components performed better when estimating the wave heading. This information was then combined with the values of the PSD total power and vessel forward speed to use as input to the NNs for SSE.
        
        A traditional NN has several layers, where each is comprised of a number of perceptrons (nodes), each of which are associated with an activation function that takes the weighted sum of the previous layer of the network as their input, and produces output for the next layer~\cite{Jain1996}. In a fully-connected NN, each perceptron in one layer is connected to every perceptron in the next layer. Weights applied to individual connections within the layers are then modified to converge on optimal outputs from the NN overall. Equation~\ref{eq:perceptron} defines the activation function of a single perceptron, where $x_{p_j}$ represents the inputs, $w_{p_j}$ are the connection weights, and $b_p$ is the bias.
        \begin{equation} \label{eq:perceptron}
            a_p = f\left( \sum_j w_{p_j} x_{p_j} + b_p \right)
        \end{equation}
        
        A multi-layer perceptron (MLP) is a simple feedforward NN, a form of supervised ML which can be used for both regression and classification problems. An MLP was used as the NN for SSE because it enables feature extraction to be completed before being input into the model, requiring a less complex architecture to define the relationship between input and output data. In this case, the NNs take response data and vessel speed as input and produce a wave property ($H_s$, $T_1$, or $\mu_h$) as output.
    
\begin{figure}[H]
            \centering
            \includegraphics[scale=0.62]{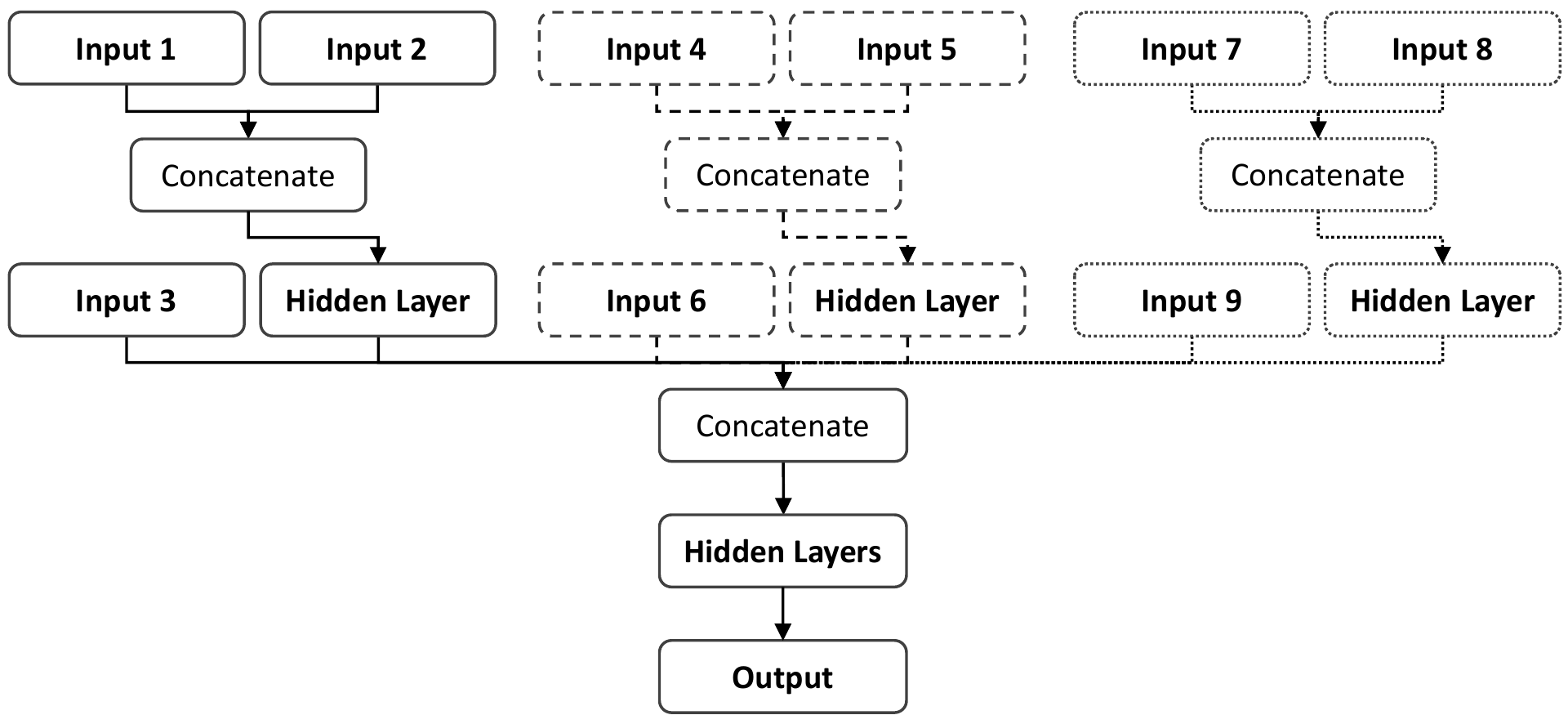}
            \caption{NN structure for 1-DOF (solid line), 2-DOF (dashed line), and 3-DOF (round-dot line) data.}
            \label{fig:nn_schem_3dof}
        \end{figure}

        Separate NNs were designed to input 1-DOF, 2-DOF, and 3-DOF responses. Figure~\ref{fig:nn_schem_3dof} shows the structure of the three different network types. The 1-DOF NN structure is outlined using a solid line, where \textit{input 1} and \textit{input 2} were defined for the spectral ordinates and associated frequency components, while \textit{input 3} contains the vessel speed and response PSD total power. The PSD ordinates and frequencies were concatenated, forming the first \textit{branch} of the NN, then passed to the first hidden layer. The output from this single hidden layer was then concatenated with \textit{input 3}, together comprising the NN \textit{trunk}, which was then passed to another hidden layer, before outputting one of the sea state properties in the final layer.
        
        The 2-DOF network is comprised of both the 1-DOF structure and the second branch of the model, comprised of \textit{input 4}, \textit{input 5}, and \textit{input 6}, outlined with a dashed line. The PSD ordinates and frequencies for a second DOF were concatenated then passed to a hidden layer, before being concatenated with the total power of the PSD, vessel speed, and data from the first DOF and passed through the network trunk. Similarly, the 3-DOF model is comprised of the 1-DOF and 2-DOF structures, as well as the third branch outlined with a dotted line and comprised of \textit{input 7}, \textit{input 8}, and \textit{input 9}, then passed through the same processes as the 1-DOF and 2-DOF models.
        
        For example, a 1-DOF model would take either heave, pitch, or roll response data as input, then output the significant wave height, mean wave period, or wave heading. The 2-DOF variant would then take heave and pitch, for example, and 3-DOF design would take all three DOFs.
        
        The \textit{rectified linear unit} activation function (ReLu) was used for each layer, while the \textit{Adam} optimizer was used to find the best set of node connection weights to map the input data to the desired output, and a \textit{mean square error} loss function was implemented to improve the model fitness.
        
        Separate NNs were trained for each combination of response DOF and sea state property ($H_s$, $T_1$, and $\mu_h$) in order to investigate the sea state information which could be extracted from the different response components.
        
        All of the NN configurations are given in~\ref{ap:A}. Following experimentation, it was found that network architectures for $H_s$ and $T_1$ estimation were optimal using the same configurations for the 1-DOF, 2-DOF, and 3-DOF models (given in Table~\ref{tab:nn_param_hs_t1}), while the network architectures for $\mu_h$ were optimal using an alternative configuration (seen in Table~\ref{tab:nn_param_mu}). All of the networks used a single hidden layer with 16 nodes in the branches (to combine the spectral ordinates and frequencies), as well as same number of training epochs (100) and loss rate (0.001). Generally, more network connections were required as additional DOFs were added to the NNs. Further, more nodes were necessary for the NNs used to estimate $\mu_h$, likely due to the fact that more PSD components were used as input to the $\mu_h$ networks, so a greater number of connections could better capture the extra information.
        
        It was chosen to perform \textit{k-fold cross-validation} to ensure there was no (or minimal) overfitting or selection bias in model~\cite{Cawley2010}. K-fold cross-validation splits the data into two sets, training data and validation data (here a split of 80/20$\%$ was used), then \textit{folds} the data set $k_{cv}$ times (for an $80$/$20$ split, $k_{cv} = 5$). Training and validation is repeated until every sample in the data has been used exactly once in the validation data set. The performance variance of the model can then be used as a metric to understand the variation in the model\textquoteright s fit, where a lower variance indicates a better fit.
        
        \subsubsection{Performance Metrics}
        
        Root mean square error (RMSE) and the coefficient of determination ($R^2$) were selected as performance metrics for the NN models. RMSE represents the sample standard deviation of the difference between the predicted and actual values, where a higher weighting is placed on greater errors, and its units are the same as the predicted variable\textquoteright s units~\cite{Jones2018}. $R^2$ is a metric of how well the model fits the data, where a value of 0 represents a model which does not fit the data at all, and a value of 1 represents a perfect fit~\cite{Jones2018}. Equations~\ref{eq:rmse} and \ref{eq:r_sqr} are examples of RMSE and $R^2$ for significant wave height, where $\hat{H}_s$ is the predicted height, $H_s$ is the actual height, and $\bar{H}_s$ is the average height. The numerator in Equation~\ref{eq:rmse} calculates the prediction residuals ($e_R = \hat{H}_{s_i} - H_{s_i}$), which gives the difference between the predicted value and true value, and can be used to assess how well the data fits the model~\cite{Jones2018}.
        \begin{equation} \label{eq:rmse}
            RMSE = \sqrt{\frac{\sum (H_{s_i} - \hat{H}_{s_i})^2}{N}}
        \end{equation}
        \begin{equation} \label{eq:r_sqr}
            R^2 = \frac{\sum (\bar{H}_s - \hat{H}_{s_i})^2}{\sum (H_{s_i} - \bar{H}_s)^2}
        \end{equation}
        
        In this case, the trained models were evaluated using a set of test data, $10\%$ of the original data set (4,800 samples), split before the training and validation data sets were split (34,560 and 8,640 samples, respectively).

\section{Results and Discussion}\label{sec:res}

    \subsection{Response Component Analysis}
    
        \subsubsection{Significant Wave Height}
        
        The performance of the network configurations presented in Table~\ref{tab:nn_param_hs_t1} for the different component models when used to estimate $H_s$ are given in Table~\ref{tab:nn_perf_hs}. 
        
        The estimation performance for $H_s$ of pitch and roll alone were shown to be quite similar; however, the performance of heave was almost an order of magnitude less than both pitch and roll in terms of RMSE, with a 100 \% better (doubled) model fitness ($R^2$). Heave alone had an $R^2$ value of 0.990, implying a very close fit of the model to the data, with an average RMSE of only $58.3mm$.  When heave was combined with pitch, the NN estimation error only slightly reduced, with heave and roll showing similar performance to heave alone. Pitch and roll together were capable of improved estimation performance; however, still with an RMSE nearly double that of the `heave only' variant\textquoteright s performance. Therefore, as expected, it can be concluded that heave is the DOF most closely related to wave height for a vessel of similar size to the USV. The models were also found to have little variation between cross-validations, implying little overfitting or selection bias.
        
        \begin{table}[H]
			\begin{center}
				\caption{$H_s$ estimation RMSE average, RMSE standard deviation, and R-squared average across k-folds for different vessel DOFs.}
				\label{tab:nn_perf_hs}
				\begin{tabular}{l | c | c | c}     %{ m|n|n } %{lll}
					Response(s)	& RMSE Avg. & RMSE Std. & R-Squared Avg. \\ \hline \hline
					Heave & $50.8mm$ & $1.92mm$ & 0.988 \\
					Pitch & $385mm$ & $4.64mm$ & 0.485 \\
					Roll & $384mm$ & $7.43mm$ & 0.486 \\
					Heave + Pitch & $55.1mm$ & $0.97mm$ & 0.990 \\
					Heave + Roll & $58.5mm$ & $0.70mm$ & 0.988 \\
					Pitch + Roll & $101mm$ & $5.05mm$ & 0.966 \\
				\end{tabular}
			\end{center}
		\end{table}

		When considering the residual error distributions for each component NN model for significant wave height estimation (see Figure~\ref{fig:comp_resid_hs}), the errors associated with the models using heave were found to be slightly worse for greater values of $\hat{H}_s$, which was also present for the pitch and roll model (although with greater spread). This implies that the models tended to underestimate $H_s$. The 1-DOF pitch and roll models both saw a steady increase in error as their $H_s$ predictions increased, before reducing again, showing the greatest errors associated with $H_s$ estimations between $1m$ and $2m$, indicating that the models were underestimating moderate seas, and overestimating smaller seas.
        
        \subsubsection{Mean Wave Period}
        
        Table~\ref{tab:nn_perf_t1} presents the results for $T_1$ estimations using the network architectures given in Table~\ref{tab:nn_param_hs_t1}. The NN models for all of the response components were found to perform well, with less variation found between the NNs than for $H_s$. Heave was shown to be the single DOF which could be used to extract the most information to estimate $T_1$, with pitch being second. The combination of both heave and pitch had the lowest errors and best model fit, with an average RMSE of $0.228s$ and an $R^2$ value of 0.991. The models all had low RMSE standard deviations, demonstrating that there was little overfitting or selection bias. 
        
        \begin{table}[H]
			\begin{center}
				\caption{$T_1$ estimation RMSE average, RMSE standard deviation, and R-squared average across k-folds for different vessel DOFs.}
				\label{tab:nn_perf_t1}
				\begin{tabular}{l | c | c | c}     %{ m|n|n } %{lll}
					Response(s)	& RMSE Avg. & RMSE Std. & R-Squared Avg. \\ \hline \hline
					Heave & $0.368s$ & $0.018s$ & 0.944 \\
					Pitch & $0.573s$ & $0.023s$ & 0.928 \\
					Roll & $0.708s$ & $0.021s$ & 0.877 \\
					Heave + Pitch & $0.228s$ & $0.022s$ & 0.991 \\
					Heave + Roll & $0.294s$ & $0.042s$ & 0.985 \\
					Pitch + Roll & $0.401s$ & $0.021s$ & 0.973 \\
				\end{tabular}
			\end{center}
		\end{table}

        The residual distributions for the different mean wave period estimation models (given in Figure~\ref{fig:comp_resid_t1} in~\ref{ap:B}) all had similar shapes, except those without heave which had a greater spread. As expected, the model for heave and pitch showed the least spread with few outliers, and the model for heave and roll showed the second best performance. Greater errors were associated with estimates closer to the centre of the $T_1$ range than the limits of the range for the roll-only model, which indicates that the NNs underestimated mean periods for moderate seas and underestimated in smaller seas (as seen previously in the results for estimated wave height $\hat{H}_s$).
        
        \subsubsection{Wave Heading}
        
	    The performance results for each of the $\mu_{h}$ component models (described in Table~\ref{tab:nn_param_mu}) are given in Table~\ref{tab:nn_perf_mu180}. The models were able to estimate wave headings with relatively high accuracy. Heave alone performed the worst of all models trained, achieving an average RMSE of $18.90^\circ$ and $R^2$ of 0.852 (averaged across the 5 k-folds). Roll also performed similarly. However, when roll was paired with pitch, the best estimation results were achieved, with an average RMSE of only $8.40^\circ$ and $R^2$ of 0.973. The standard deviations of the RMSE varied considerably, with heave alone showing the greatest selection bias and/or overfitting.
	    
	    \begin{table}[H]
			\begin{center}
				\caption{$\mu_h$ estimation RMSE average, RMSE standard deviation, and R-squared average across k-folds for different vessel DOFs.}
				\label{tab:nn_perf_mu180}
				\begin{tabular}{l | c | c | c}     %{ m|n|n } %{lll}
					Response(s)	& RMSE Avg. & RMSE Std. & R-Squared Avg. \\ \hline \hline
					Heave & $18.9^\circ$ & $1.13^\circ$ & 0.852 \\
					Pitch & $10.5^\circ$ & $0.195^\circ$ & 0.956 \\
					Roll & $18.3^\circ$ & $0.732^\circ$ & 0.866 \\
					Heave + Pitch & $8.88^\circ$ & $0.473^\circ$ & 0.970 \\
					Heave + Roll & $12.8^\circ$ & $0.157^\circ$ & 0.935 \\
					Pitch + Roll & $8.40^\circ$ & $0.221^\circ$ & 0.973 \\
				\end{tabular}
			\end{center}
		\end{table}
		
		A number of patterns can be seen in the residual distributions for the wave heading estimations for each component model (shown in Figure~\ref{fig:comp_resid_mu180}). For heave alone, the greatest errors are generally estimated around beam sea. This is similar to what was seen for the significant wave height and to a lesser extent mean wave period estimations, with the limits of the wave property range being underestimated and overestimated. However, the greatest outlying errors for most of the models increased in magnitude as they approach head and following sea estimations. This shows that at times the models would estimate head seas as following seas, and vice versa. This can be explained by the similar motion responses induced by the same wave input using the transfer functions for a vessel in head and following seas for both pitch and roll.

    \subsection{3-DOF Models}
    
    Finally, the three vessel response DOFs, heave, pitch, and roll, were combined into a single NN to estimate the wave conditions.
	
	The results from the k-fold cross-validation for the 3-DOF models are given in Table~\ref{tab:nn_perf_3dof}. As expected, the inclusion of all of the data led to best performance of the models.

	\begin{table}[H]
		\begin{center}
			\caption{3-DOF  NN model performance.}
			\label{tab:nn_perf_3dof}
			\begin{tabular}{l | c | c | c }
				Metric	& $H_s$ & $T_1$ & $\mu_h$ \\ \hline \hline
				RMSE Avg. & $53.8mm$ & $0.171s$ & $7.21^\circ$ \\
				RMSE Std. & $1.16mm$ & $0.006s$ & $0.411^\circ$ \\
				R-squared & 0.990 & 0.995 & 0.980 \\
			\end{tabular}
		\end{center}
	\end{table}
	
	The 3-DOF $H_s$ model did not see a large increase in performance when compared with the heave and pitch model. The average RMSE reduced to $53.8mm$, with a standard deviation of $1.16mm$, and an average $R^2$ value of 0.990. The similar performance could indicate that similar information was extracted from pitch and roll to estimate $H_s$.
	
	The $T_1$ estimation performance was, however, improved more considerably when using all of the vessel response data, with an average RMSE of $0.171s$ and standard deviation of $0.006s$, and an average $R^2$ of 0.995. The closest performance was achieved with the model trained on the heave and pitch data. This indicates that the information extracted from each DOF was considerably different from each other when responding to varying $T_1$.
	
	The $\mu_{h}$ estimator experienced some improvement in estimation performance when heave was included, compared to only pitch and roll, but with a lower average RMSE and slightly greater $R^2$ value. However, this came at the cost of a greater RMSE standard deviation, meaning that there was a greater variation in model performances between k-fold validations.
	
\begin{figure}[H]
        \centering
        \includegraphics[scale=0.8]{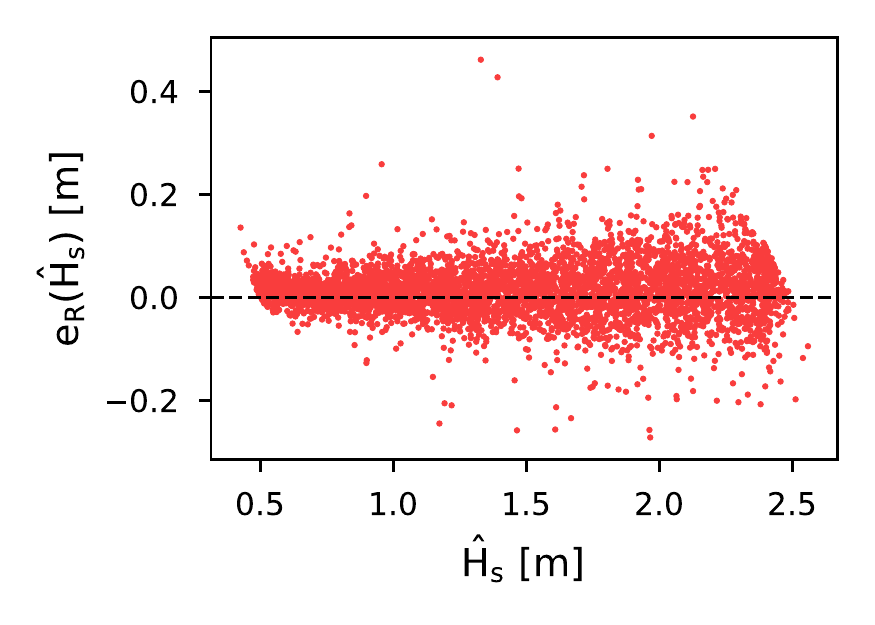}
        \caption{$H_s$ estimations versus residual errors for 3-DOF NN model.}
        \label{fig:3dof_resid_hs}
    \end{figure}

\begin{figure}[H]
        \centering
        \includegraphics[scale=0.8]{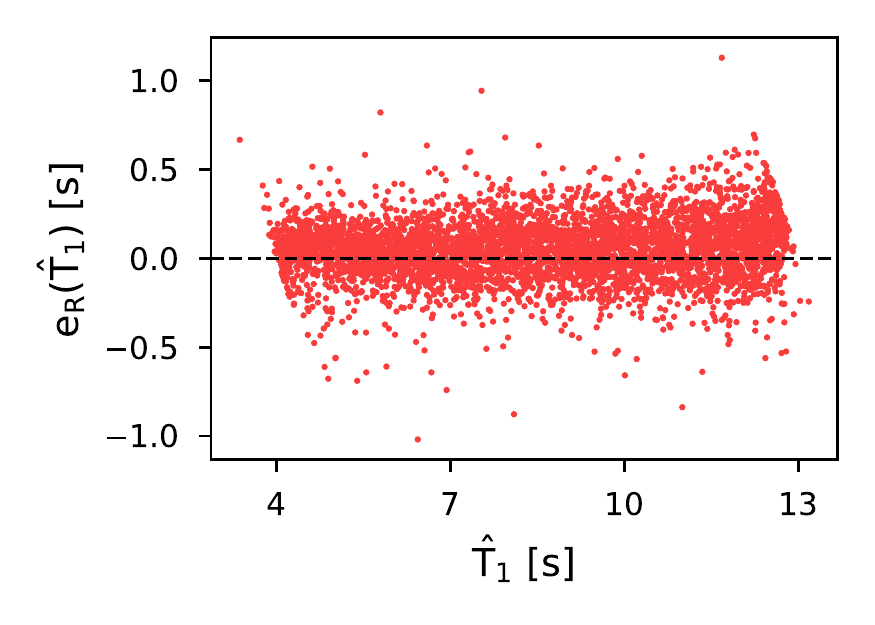}
        \caption{$T_1$ estimations versus residual errors for 3-DOF NN model.}
        \label{fig:3dof_resid_t1}
    \end{figure}

\begin{figure}[H]
        \centering
        \includegraphics[scale=0.8]{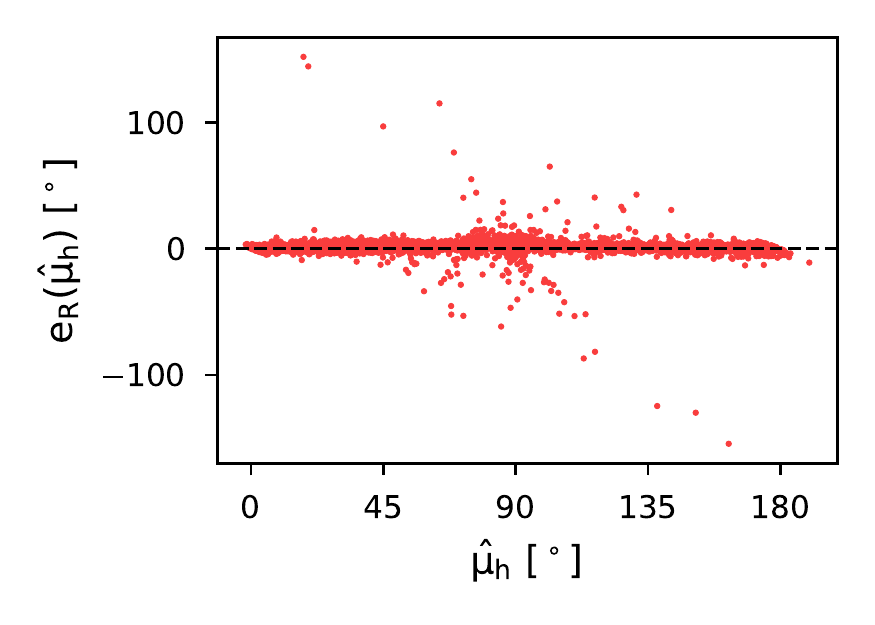}
        \caption{$\mu_h$ estimations versus residual errors for 3-DOF NN model.}
        \label{fig:3dof_resid_mu}
    \end{figure}

	The residual plots for the 3-DOF wave estimation NNs are presented in Figures~\ref{fig:3dof_resid_hs} to~\ref{fig:3dof_resid_mu}. Each of the models show similar patterns to those of the best performing 2-DOF models.
	
    Figure~\ref{fig:3dof_resid_hs} demonstrates a general trend towards larger error for greater $H_s$ estimations, while the 3-DOF $T_1$ residuals in Figure~\ref{fig:3dof_resid_t1} show that the model is biased towards underestimating $T_1$ values. 
	
    In Figure~\ref{fig:3dof_resid_mu} it can be seen that the estimations have the greatest accumulation of errors about beam sea. The model is, therefore, tending to incorrectly estimate the vessel as being in beam seas more frequently than in head or following seas. As seen with the $\mu_h$ component model estimations, the outlying greatest errors are seen closer to head and following sea, with the model occasionally mistaking the vessel as being in opposing wave headings.
    
    As a point of comparison with one another, the RMSE for $\mu_h$ as a ratio of the range of $\mu_h$ values tested was $4.01\%$. The ratio of RMSE to range for $H_s$ was $2.69\%$, and the ratio for $T_1$ was $1.90\%$. Therefore, the relative estimation performance was best for $T_1$ and worst for $\mu_h$. However, a comparison with results obtained from other studies provides a better insight into the relative performance of the SAWB-inspired SSE procedure presented here.
    
    \subsection{Low-power Responses}\label{sec:lpr}
    
    In order to investigate the relationship between vessel motion response magnitude and SSE performance, the total power ($m_0$) in the heave, pitch, and roll response spectra were plotted against absolute estimation errors ($e_A$) for $H_s$, $T_1$, and $\mu_h$ estimation. It was assumed that the magnitude of the responses are related to the wave frequencies and height, where higher frequencies and lower heights would result in lower total power in the responses.
    
\begin{figure}[H]
        \centering
        \includegraphics[scale=0.8]{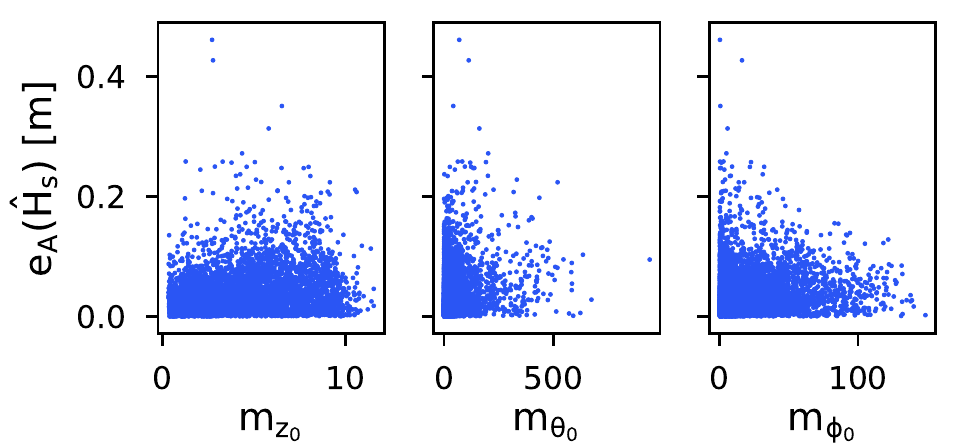}
        \caption{Total power in heave, pitch, and roll response versus absolute $\hat{H}_s$ error.}
        \label{fig:hs_m0}
    \end{figure}

    Figure~\ref{fig:hs_m0} shows the total power of heave, pitch, and roll responses versus absolute $H_s$ estimation error. The estimation accuracy appears to get worse as the total power reduces for the pitch and roll responses; however, this relationship does not appear to be present for the heave responses, showing that the small vessel was still able to capture sufficient information to estimate $H_s$ when in high-frequency, low wave height sea states.
    
\begin{figure}[H]
        \centering
        \includegraphics[scale=0.8]{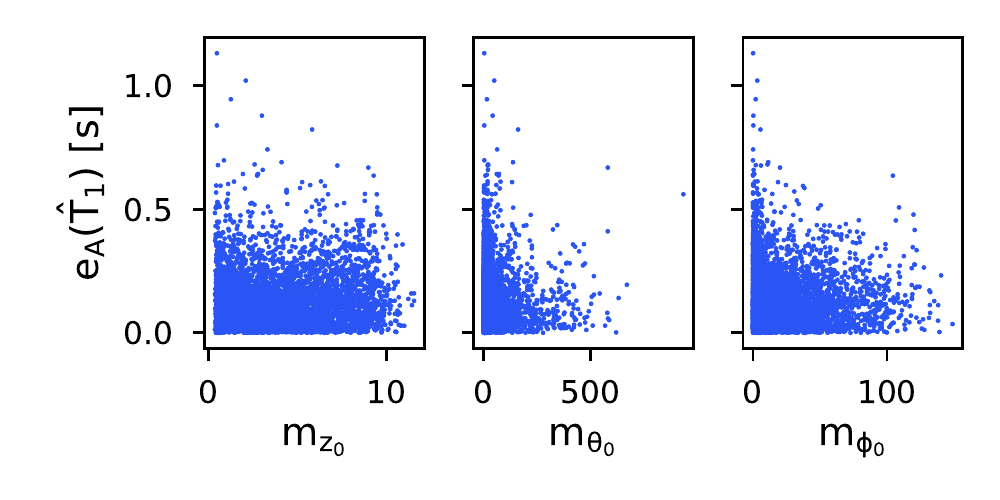}
        \caption{Total power in heave, pitch, and roll response versus absolute $\hat{T}_1$ error.}
        \label{fig:t1_m0}
    \end{figure}

    The general trend of residual errors in Figure~\ref{fig:t1_m0} were similar Figure~\ref{fig:hs_m0}, where it can be seen that the $T_1$ estimation performance reduces as the total power of the pitch and roll responses decrease. When considering the heave response, this relationship was much less pronounced, with the magnitude of errors being approximately evenly distributed across all total heave response powers.
    
    The $\mu_h$ estimation performance decreases considerably in conditions associated with low pitch and roll response, where a greater concentration of large errors are associated with low responses. The same relationship between greater errors and low heave responses can be seen; however, the relationship is less pronounced and large errors were distributed across the heave power range.
    
    \begin{figure}[H]
        \centering
        \includegraphics[scale=0.8]{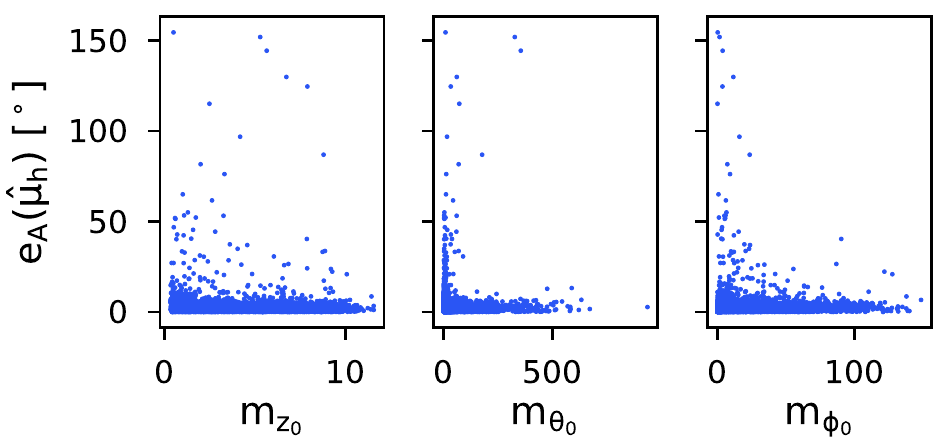}
        \caption{Total power in heave, pitch, and roll response versus absolute $\hat{\mu}_{h}$ error.}
        \label{fig:mu_m0}
    \end{figure}
    
    The results show that while there is a trend towards decreased SSE accuracy for lower power motion responses of the USV, heave responses are still able to estimate $H_s$ and $T_1$ with similar accuracy across all total power levels, while $\mu_h$ estimation seems most affected by the motion response magnitude of the vessel.
    
        \subsubsection{Results Comparison}
        
        A wide variety of SSE studies using SAWB have been undertaken, where results for the SSEs are presented in many different ways. In order to critically evaluate the results from the described SSE approach, two previous studies providing similar results metrics have been selected for direct comparison.
        
        The study by Arneson, Brodtkorb, and S{\o}rensen~\cite{Arneson2019} is one of the only other ML-based SAWB approaches that uses spectral data to train SSE models (rather than time-series data), and uses similar metrics to this study for presenting the results. The authors simulated their wave input using a parametric JONSWAP spectrum, using QDA to estimate a wave heading classification and PLSR to then estimate significant wave height and peak wave period. Differences in the approach presented in this paper from the one presented in~\cite{Arneson2019} include analysis of the cross-spectra, resulting in full $360^\circ$ wave heading estimation, simulating rougher sea states, and 6-DOF responses were used as input. Further, Arneson, Brodtkorb, and S{\o}rensen only used the total power of response spectra as SSE model input. Therefore, their model has been compared using only their results for comparable sea states, while their wave heading angle errors have been compared to the equivalent wave heading errors found in this study. Further, the errors associated with peak wave period were compared with the errors associated with mean wave period using the designed approach.
        
        Hinostroza and Soares~\cite{Hinostroza2016} developed a model-based approach to SSE using SAWB, which estimated significant wave height and peak wave period for a parametric JONSWAP spectrum, as well as wave heading (within a $180^\circ$ range). The authors utilized genetic algorithms to calibrate inverse transfer functions which mapped the heave, pitch, and roll responses to the wave spectrum. Hinostroza and Soares~\cite{Hinostroza2016} tested their method using two different ship types via simulation, as well as via experimentation. The simulation results from the smaller vessel are compared with the developed methodology, while 2 out of 8 cases were omitted as they simulated rough sea states outside the range used for this study.
        
        Table~\ref{tab:sse_nn_comp} compares the findings from~\cite{Arneson2019} and~\cite{Hinostroza2016} with the developed 3-DOF SSE NNs. 
        
        \begin{table}[H]
    		\begin{center}
    			\caption{SSE NN performance comparison.}
    			\label{tab:sse_nn_comp}
    			\begin{tabular}{l | c | c | c}
    				Study	& Sig. Wave Height & Wave Period & Wave Heading \\
    				& RMSE & RMSE & RMSE \\ \hline \hline
    				Long et al. 2022 & 53.8$mm$ & 0.171$s$ & 7.21$^\circ$ \\ \hline
    				Arneson et al.~\cite{Arneson2019} & 60$mm$ & 1.5$s$ & 4$^\circ$ \\ \hline
    				Hinostroza and  & 23$mm$ & 1.1$s$ & 9$^\circ$ \\ 
    				Soares~\cite{Hinostroza2016} & $\ $ & $\ $ & $\ $ \\
    			\end{tabular}
    		\end{center}
	    \end{table}
        
        When examining the RMSE values for mean wave period, it can be seen that the methodology designed in this work achieved a result that was almost an order of magnitude more accurate than Arneson, Brodtkorb, and S{\o}rensen\textquoteright s~\cite{Arneson2019}, while the RMSE of Hinostroza and Soares~\cite{Hinostroza2016} was more than six times greater than for this study. The significant wave height estimations were similar to those reported in~\cite{Arneson2019}, while RMSE was more than halved compared to the results presented in~\cite{Hinostroza2016}. The results for estimated wave headings were found to be more accurate using Arneson, Brodtkorb, and S{\o}rensen\textquoteright s classification method than the proposed method, while Hinostroza and Soares\textquoteright s methodology showed slightly worse performance. It should be noted that the results from the work reported here comprised of 4800 instances, while there were only five comparable results from~\cite{Arneson2019} and six from~\cite{Hinostroza2016}. Further, the relative ship geometries used in the compared SSE methodologies~\cite{Arneson2019,Hinostroza2016} could have an effect on the estimation performance. 
        
        Comparing the estimation accuracy of a small craft with low responses to that of the 14,000TEU container ship presented in Kawai et al.\textquoteright s study~\cite{Kawai2021}, the same decrease in estimation performance at low responses cannot be seen for the small USV model presented in Section~\ref{sec:lpr}. $H_s$ and $T_1$ estimation saw no major decrease in accuracy when heave was considered, while trends for each of the motion response components were not as great as those in~\cite{Kawai2021}. The presented results emphasise the relationship between vessel size and SSE performance as a function of wave properties when using the SAWB technique.
        
        Overall, the NN models for SSE designed during this research show promising results when compared with similar approaches. With further development, there is potential for the described methodology to be expanded upon for performance augmentation and possible future implementation in physical systems.

\section{Conclusion}\label{sec:concl}

A methodology for estimating sea states using a small USV has been developed, which harnesses the benefits of ML as an alternative to fine tuning physics-based models. Artificial neural network models trained using heave, pitch, and roll vessel response data have been shown to be able to estimate significant wave heights, mean wave periods, and relative wave headings effectively for idealized sea states (within the given constraints). The information which could be extracted from each motion response has been analysed, with results showing a strong correlation between heave responses and significant wave height estimates, whilst the accuracy of mean wave period and wave heading predictions were observed to improve considerably when data from multiple vessel degrees of freedom was utilized. Mean wave period was estimated with the highest accuracy, while wave heading was estimated with the lowest accuracy, with pitch found to be the motion component which best estimated wave heading. Overall, the neural networks trained using 3-degree-of-freedom (heave, pitch and roll) motion for sea state estimation were shown to perform well when compared to existing approaches that use similar simulation setups. Finally, the relationship between USV motion response power and estimation accuracy was explored and compared to a similar approach with a larger vessel. Results showed that the smaller vessel could achieve relatively higher accuracy estimations for lower vessel response magnitudes, with heave shown to capture the most information for sea states with high frequencies and low wave heights.

In the future, a number of extensions could be made to this study. Firstly, more complex vessel dynamic and transfer function models should be used which do not assume decoupling between DOFs so that the cross-spectra can be analysed to differentiate between port and starboard wave headings. Secondly, the networks could be evaluated using model-scale testing data under controlled conditions. Thirdly, the PSD analysis procedure could be further optimized and experimented with to improve SSE performance. Finally, NNs can be tested for a greater range of conditions by using a simulation model capable of capturing non-linear behaviours associated with rougher sea states, short crested seas, and combined wind-swell conditions, thereby improving the robustness of the model and estimations.

% \section*{References}

\bibliography{AOR_Article}

\newpage

\appendix

\section{Neural Network Configurations}~\label{ap:A}

\begin{table}[H]
	\begin{center}
		\caption{$H_s$ and $T_1$ NN model configurations.}
		\label{tab:nn_param_hs_t1}
		\begin{tabular}{l | c | c | c }
			Element	& 1-DOF &  2-DOF & 3-DOF \\ \hline \hline
			Branch hidden layers & 1 & 1 & 1 \\
			Branch nodes & [16] & [16] & [16] \\
			Trunk hidden layers & 2 & 3 & 3 \\
			Trunk nodes & [16,8] & [16,8,8] & [32,32,16] \\
			Batch size & 32 & 16 & 16 \\
			Epochs & 100 & 100 & 100 \\
			Loss rate & 0.001 & 0.001 & 0.001 \\
		\end{tabular}
	\end{center}
\end{table}

\begin{table}[H]
	\begin{center}
		\caption{$\mu_h$ NN model configurations.}
		\label{tab:nn_param_mu}
		\begin{tabular}{l | c | c | c }
			Element	& 1-DOF &  2-DOF & 3-DOF \\ \hline \hline
			Branch hidden layers & 1 & 1 & 1 \\
			Branch nodes & [16] & [16] & [16] \\
			Trunk hidden layers & 2 & 3 & 3 \\
			Trunk nodes & [32,16] & [32,32,16] & [32,32,16] \\
			Batch size & 16 & 32 & 32 \\
			Epochs & 100 & 100 & 100 \\
			Loss rate & 0.001 & 0.001 & 0.001 \\
		\end{tabular}
	\end{center}
\end{table}

\newpage

\section{SSE Component Residual Plots}~\label{ap:B}

\begin{figure}[H]
    \centering
    \includegraphics[scale=0.7]{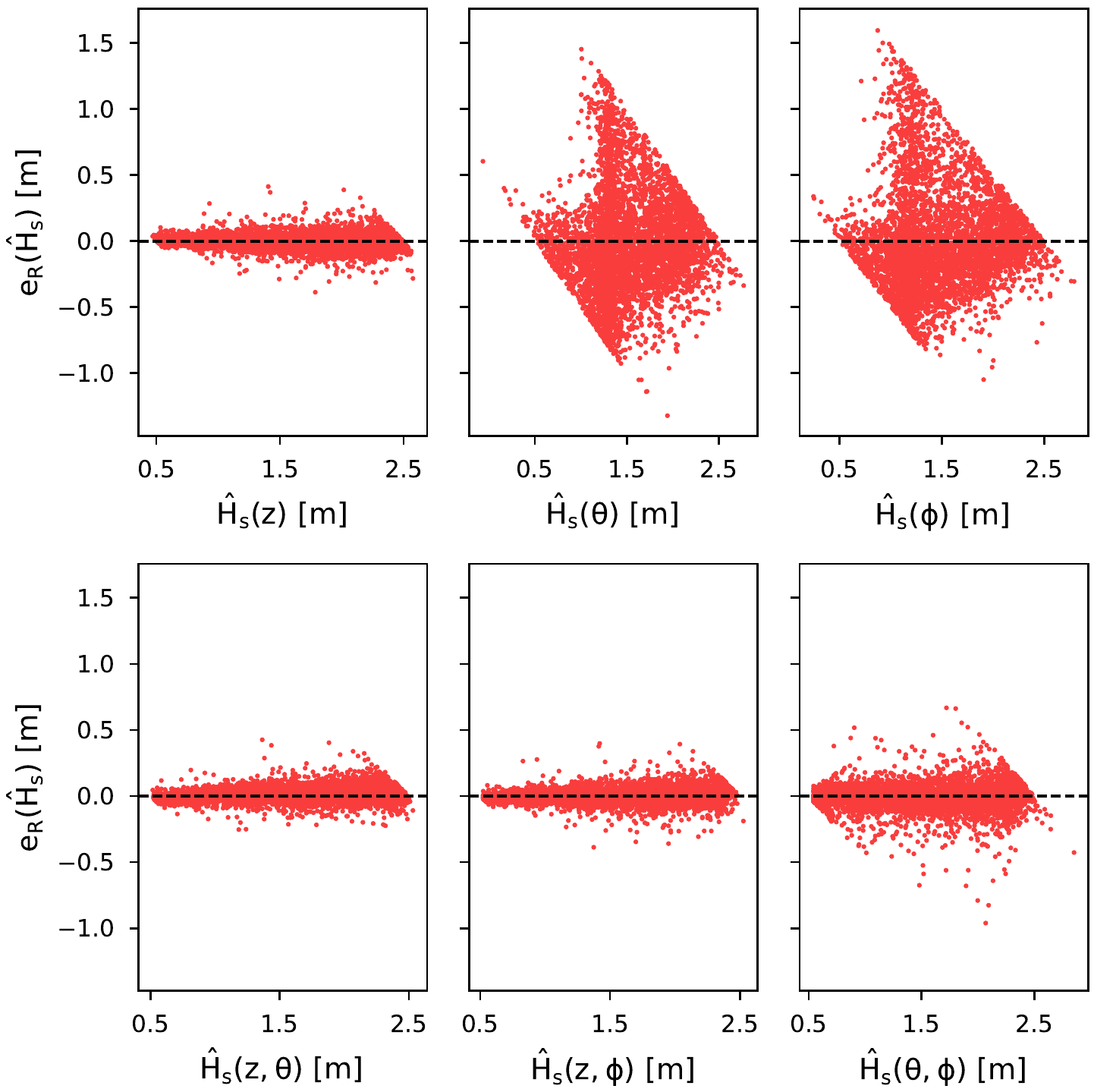}
    \caption{Residual plots for $\mathrm{H_s}$ component estimations.}
    \label{fig:comp_resid_hs}
\end{figure}

\begin{figure}[H]
    \centering
    \includegraphics[scale=0.7]{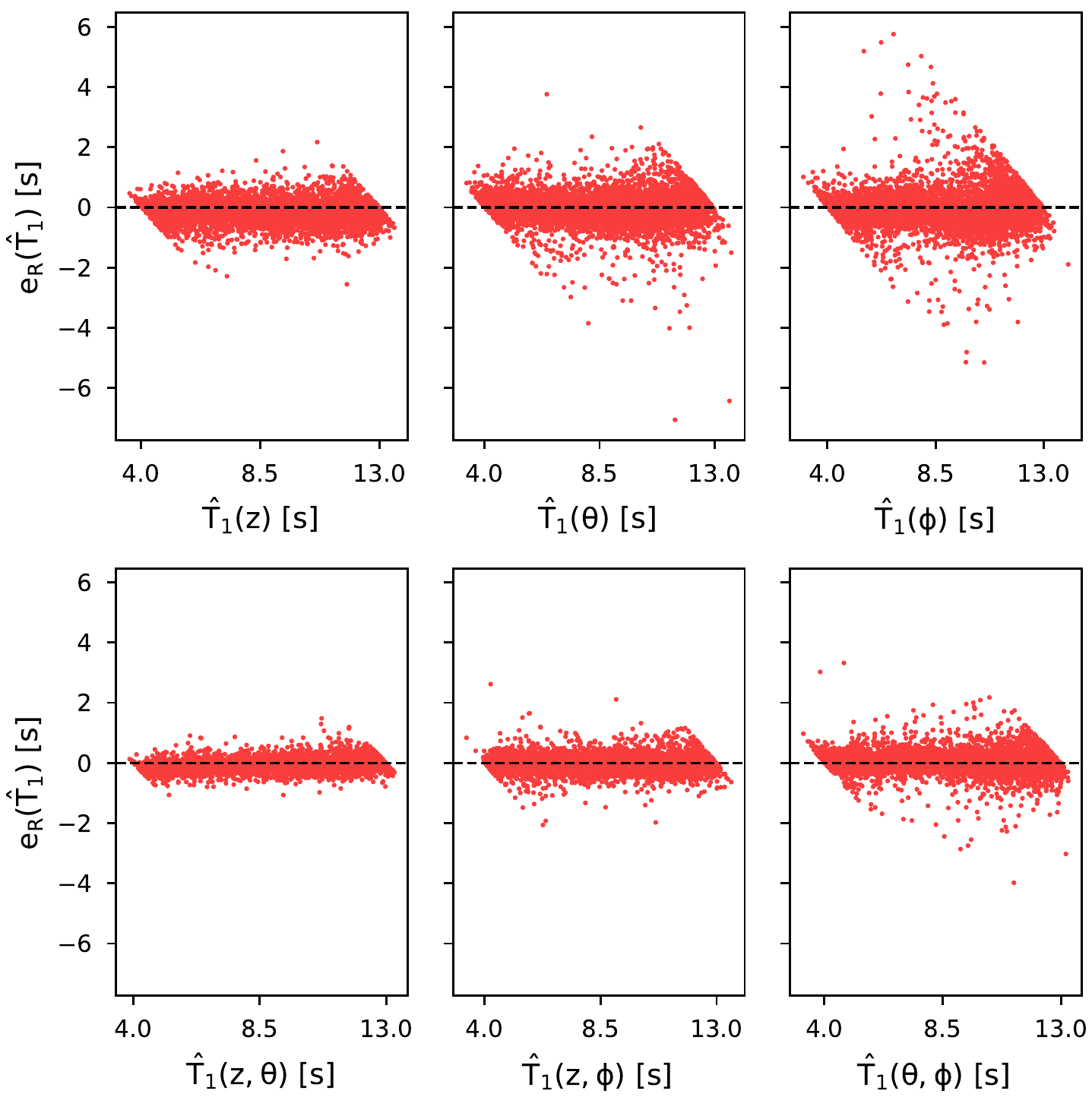}
    \caption{Residual plots for $T_1$ component estimations.}
    \label{fig:comp_resid_t1}
\end{figure}

\begin{figure}[H]
    \centering
    \includegraphics[scale=0.8]{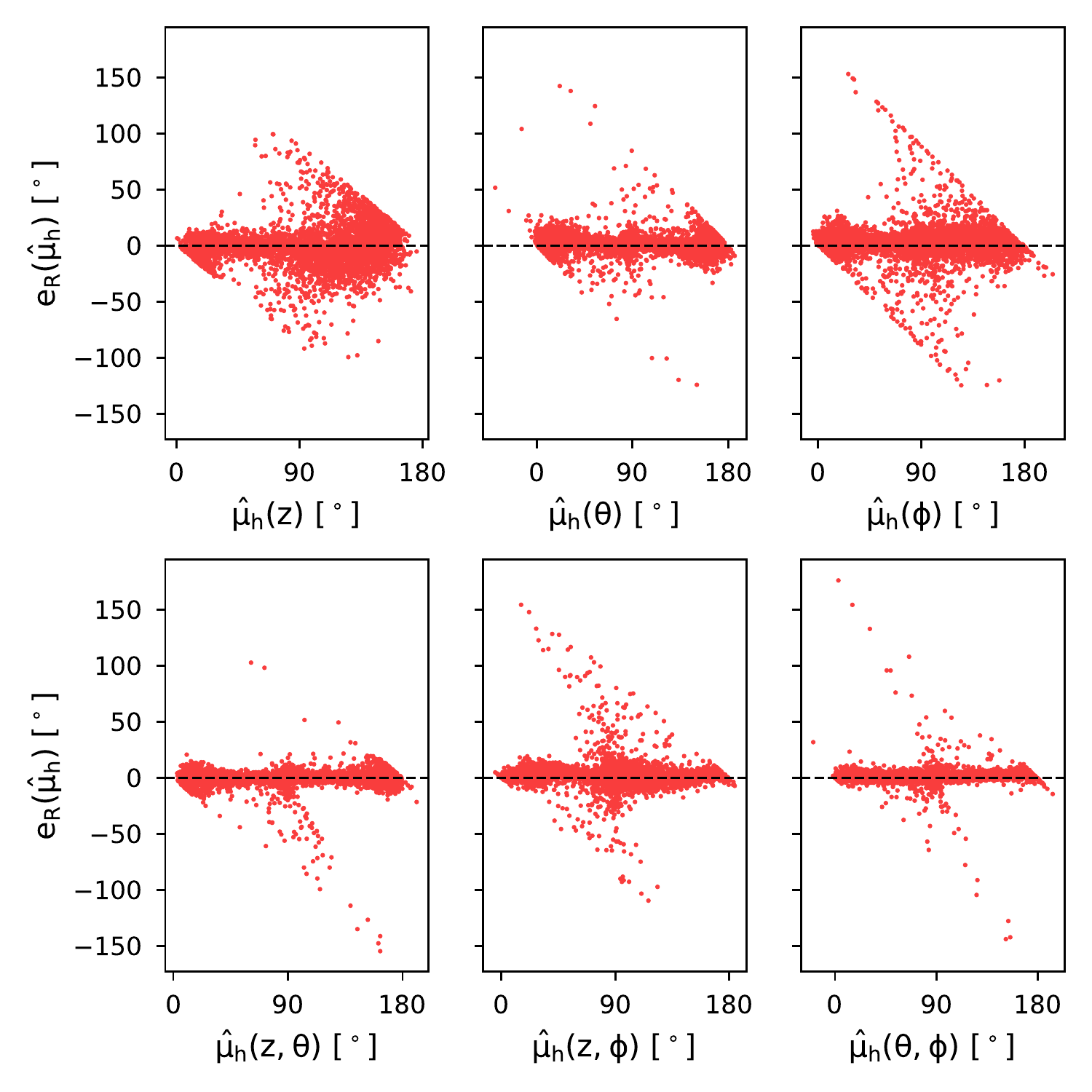}
    \caption{Residual plots for $\mathrm{\hat{\mu}_{h}}$ component estimations.}
    \label{fig:comp_resid_mu180}
\end{figure}

\end{document}